\documentclass{article}

\PassOptionsToPackage{numbers,compress}{natbib}
    \usepackage[final]{casml_2024}

\usepackage[utf8]{inputenc} 
\usepackage[T1]{fontenc}    
\usepackage[hidelinks]{hyperref}       
\usepackage{url}            
\usepackage{booktabs}       
\usepackage{amsfonts}       
\usepackage{nicefrac}       
\usepackage{microtype}      
\usepackage{xcolor}         
\usepackage{subfig}
\usepackage{graphicx}
\usepackage{amsmath}
\usepackage{lineno}

\title{TAUDiff: Highly efficient kilometer-scale downscaling using generative diffusion models}

\author{%
  Rahul Sundar \\
  Verisk Analytics \\
  Hyderabad, IND 500081 \\
  \texttt{rahul.sundar@verisk.com} \\
  \And
   Yucong Hu \\
  Verisk Analytics \\
  Ottawa, Ontario \\
   Canada K1J 9J8 \\
  \texttt{yucong.hu@verisk.com} \\
  \And
  Nishant Parashar \\
  Verisk Analytics \\
  Hyderabad, IND 500081 \\
  \texttt{nparashar@verisk.com} \\
  \And
  Antoine Blanchard \\
  Verisk Analytics \\
  Boston, MA 02115 \\
  \texttt{ablanchard@verisk.com} \\
  \And
  Boyko Dodov \\
  Verisk Analytics \\
  Boston, MA 02115 \\
  \texttt{bdodov@verisk.com} \\
}

\begin{document}
\maketitle

\begin{abstract}
Deterministic regression-based downscaling models for climate variables often suffer from spectral bias, which can be mitigated by generative models like diffusion models. To enable efficient and reliable simulation of extreme weather events, it is crucial to achieve rapid turnaround, dynamical consistency, and accurate spatio-temporal spectral recovery. We propose an efficient correction diffusion model, TAUDiff, that combines a deterministic spatio-temporal model for mean field downscaling with a smaller generative diffusion model for recovering the fine-scale stochastic features. We demonstrate the efficacy of this approach on downscaling atmospheric wind velocity fields obtained from coarse GCM simulations. We then extend TAUDiff for computationally efficient kilometer-scale downscaling of atmospheric wind velocity fields. Owing to low inference times, our approach can ensure quicker simulation of extreme events necessary for estimating associated risks and economic losses. 
\end{abstract}

\section{Introduction}
Weather extremes are on the rise due to accelerated climate change~\cite{hoeppe2016trends}. Given their potential to severely damage life and property, it is becoming increasingly important to estimate their frequency, associated risks and economic losses beforehand~\citep{houser2015economic,field2012managing,sec2022disclosure}. By insuring for such losses, we can become more resilient towards extreme events~\cite{robinson2021risk}. Climate risk modeling often relies on historical Earth system observations~\cite{hersbach2020era5} or physics-based general circulation models (GCMs)~\citep{wang2014global} to generate climate projections. Typically, GCMs operate at a coarse resolution ($O(10)-O(10^2)$km) due to compute limitations. This leads to incorrect characterization of weather extremes. In recent years, machine-learning based statistical downscaling approaches have been explored to obtain realistic well-resolved climate data over specific regions~\cite{park2022downscaling,blanchard2022multi, daust2024capturing}. These methods leverage historical Earth system observation data to create a non-linear mapping from bias-corrected coarse GCM simulations to the desired higher-resolution outputs. 

While deterministic regression models effectively capture large-scale features, they struggle with fine-scale stochastic atmospheric processes due to low-frequency spectral bias~\cite{xu2022overview}. This limitation has recently led to the adoption of generative models like GANs~\cite{daust2024capturing, rampal2024robust,li2024generative}, and denoising diffusion models for downscaling tasks~\cite{mardani2023generative, watt2024generative}. Denoising diffusion models~\cite{song2020denoising, ho2020denoising,song2020score} are particularly promising due to their stability in training, reliable convergence, and high output quality. However, sampling is often time consuming. Addressing this, Karras {\it et al.}~\cite{karras2022elucidating} explored the design space of such diffusion models, and proposed the elucidated diffusion model (EDM) which successfully reduced the number of model evaluations (from $O(10^3)$ to $O(10)$) required to generate a single sample. 
Motivated by this, a correction diffusion model (CorrDiff)~\cite{mardani2023generative} was proposed for kilometer-scale downscaling. CorrDiff combined a UNet-based deterministic model to map the mean field and an EDM correction to capture fine-scale stochastic content. More recently, Merizzi~{\it et al.}~\cite{merizzi2024wind} proposed an ensemble-diffusion model for kilometre-scale downscaling $10$-metre wind speeds. Moreover, with their ensemble-diffusion model, they reported an inference time of approximately two hours for one year worth of data.

In the context of extreme-event simulation, it is vital that both short- and long-term event statistics of downscaled data be consistent with historical observations.
So, the lack of temporal modeling in downscaling models may affect dynamical consistency of downscaled data (e.g. distorted propagation of storm fronts). One could address this issue by borrowing techniques from video generation/prediction~\cite{tan2023openstl}, as explored by Yoon {\it et al.}~\cite{yoon2023deterministic} for regional weather forecasting. However, such techniques have not yet been explored for downscaling. Moreover, large models as in ~\cite{mardani2023generative} are computationally intensive to train and infer. This prohibits the generation of even relatively small ($O(10^3)$) extreme-event datasets, which are crucial for accurately quantifying climate tail risk. Given a good mean-field model, it is possible that a smaller and computationally efficient diffusion model would suffice. This would reduce overall computational demands, inference times, and improve efficiency for real-time use. 

To address these challenges, we propose a computationally efficient \textbf{T}emporal \textbf{A}ttention \textbf{U}nit enhanced \textbf{Diff}usion model (TAUDiff) that integrates (a) a video prediction model for dynamically consistent mean-field downscaling, and (b) a smaller guided denoising diffusion model for stochastically generating the fine-scale features. We train the models on atmospheric wind fields obtained from reanalysis dataset. The performance of TAUDiff is first compared against separate mean-field regression and end-to-end diffusion models under a fixed training budget. We evaluate the downscaling performance of TAUDiff on an 
ensemble of bias-corrected coarse GCM outputs using various spectral statistics. As a proof of concept, we also evaluate if TAUDiff is capable of mapping a reanalysis dataset of $0.25^\circ$ (about 25km) grid resolution to another dataset at $0.0625^\circ$ (about 5km) grid resolution.
We finally discuss its potential in producing accurate and computationally efficient extreme-event datasets, and the implications of model inference times and carbon footprint offsetting.

\section{Methods}
\paragraph{Overview} We demonstrate the efficacy of TAUDiff in downscaling atmospheric wind fields over the European region through two case studies: (i) model performance validation and GCM output downscaling to $0.25^{\circ}$ resolution, and (ii) extension of TAUDiff to kilometer-scale downscaling ($0.25^{\circ}$ to $0.0625^{\circ})$. Instead of a single time instance input like in~\cite{mardani2023generative}, our approach uses a deterministic mean-field regression component that takes a temporal sequence of coarse wind velocity snapshots with orography data as input. Here, the high-resolution wind fields from the final time step of the sequence serves as the target. We then train a generative diffusion model for correcting the outputs of the mean-model. We discuss the mean-field and diffusion model components of TAUDiff below. 

\begin{figure}[!htbp]
    \centering
    \includegraphics[width=\linewidth]{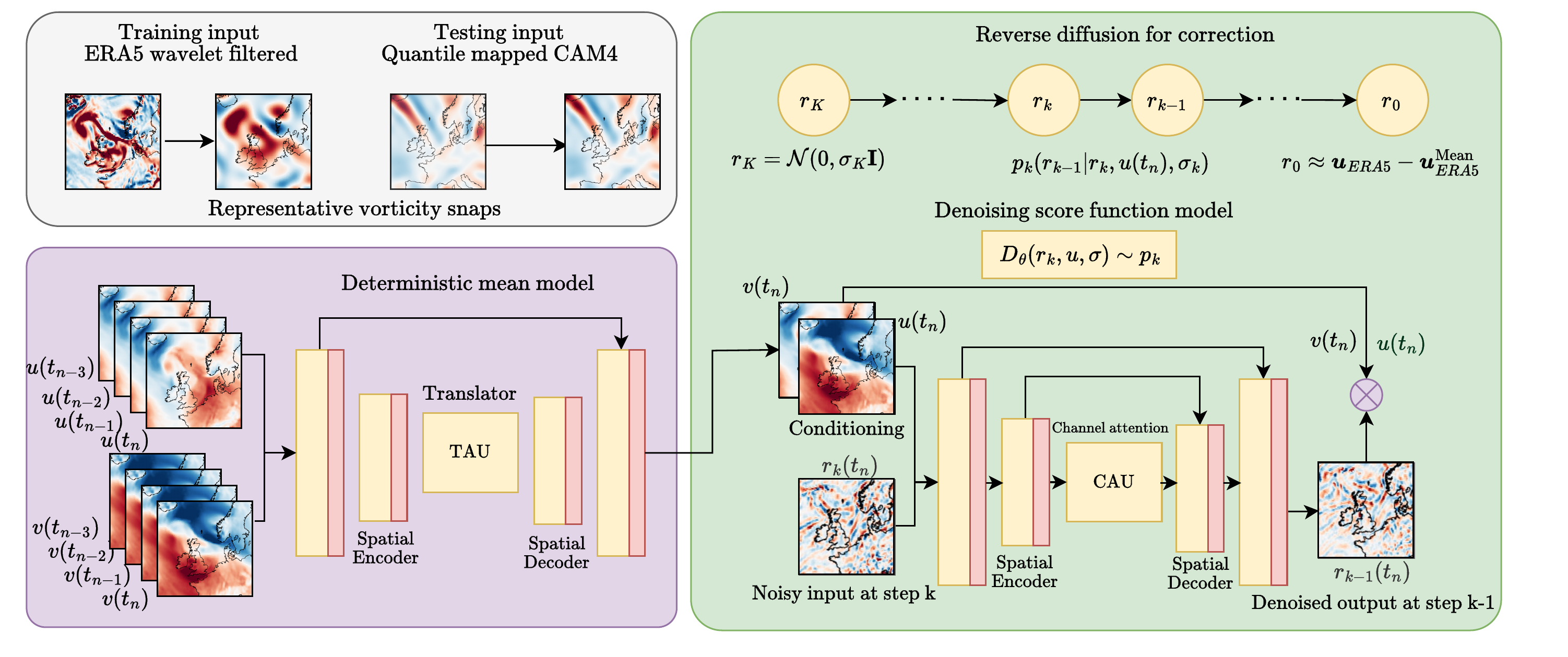}
    \caption{Schematic of the TAUDiff model}
    \label{fig:schematic}
\end{figure}

\paragraph{TAUDiff Component 1: Mean model}
We adopt the Simple yet better Video Prediction  (SimVP) architecture~\cite{gao2022simvp} consisting of a spatial backbone, and a translator for temporal modelling, ensuring temporal coherence and simplicity as compared to the more complex transformer based architectures~\cite{tan2023openstl}. We specifically consider a UNet for the spatial backbone, and the temporal attention unit (TAU)~\cite{tan2023temporal} for the translator. The TAU first independently models spatial dependency via static, and both cross-channel and temporal dependencies using dynamical attention units, respectively, and then combines them. We train the mean model using a weighted combination of mean absolute error (MAE), mean squared error (MSE), and also to maintain dynamical consistency, physics-based losses on advection ($\boldsymbol{u}\cdot\nabla\boldsymbol{u}$), vorticity ($\nabla\times\boldsymbol{u}$) and divergence ($\nabla\cdot\boldsymbol{u}$) of wind fields ($\boldsymbol{u}$) are considered. 
Although dynamically consistent predictions are possible with this mean model, the downscaled fields still lack the stochastic fine scale features. This is where Component 2 comes into play. 

\paragraph{TAUDiff Component 2: Correction diffusion model}
To capture the residual stochastic fine scale features (which cannot be captured by the mean model), we build a relatively small correction diffusion model ($\sim O(1)$ million (M) parameters) trained using a score-matching loss~\cite{song2020improved}. To maintain consistency of our approach, we use a SimVP architecture as in the mean model but with a residual dense UNet as the spatial backbone.
Once the model is trained, a data sample can be generated by solving a stochastic differential equation modelling a reverse diffusion process~\cite{ho2020denoising, karras2022elucidating}. Since the conditional input to the diffusion model is the mean model output (for a single time instance), the TAU morphs into a Channel Attention Unit (CAU) where the dynamical attention unit now models cross-channel dependencies and their relative importance (see figure~\ref{fig:schematic} for a detailed schematic of TAUDiff.).

\section{Results and discussion}
\subsection{Model validation and GCM downscaling}
\begin{figure}[!htbp]
    \centering
    \includegraphics[width=\linewidth]{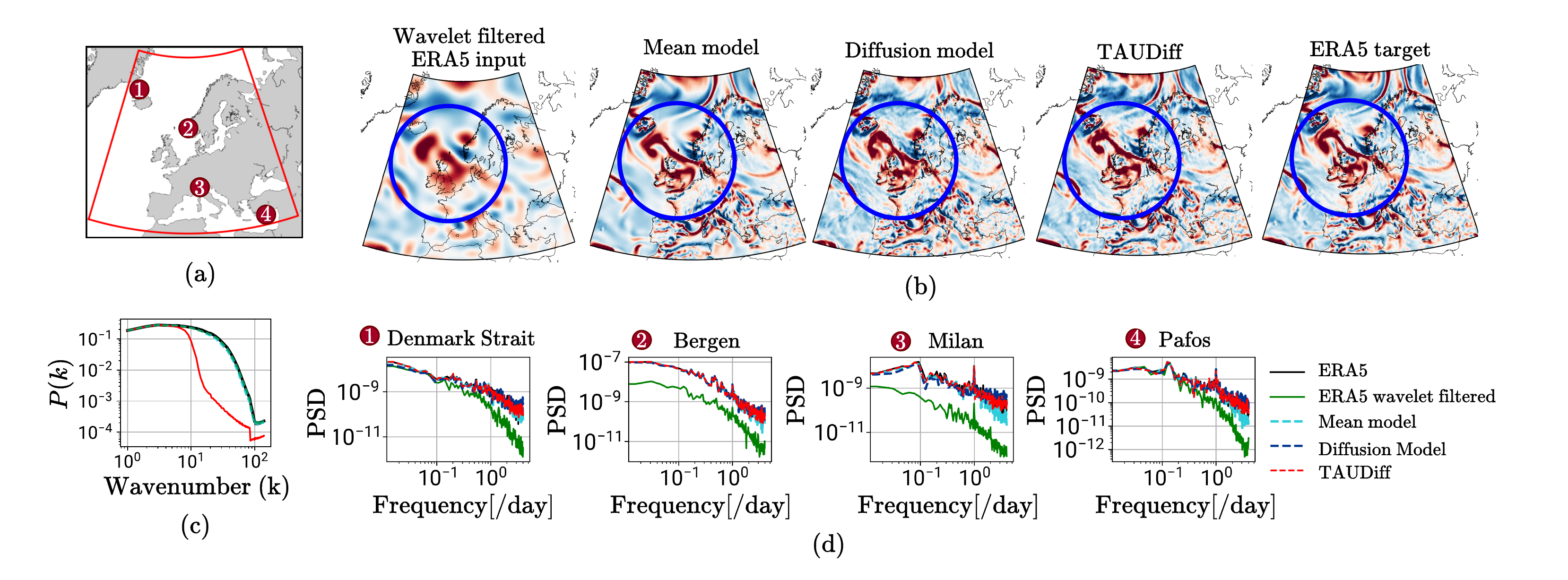}
    \caption{(a) European region used for training the downscaling models, with select locations used for evaluating performance. Comparison of model predictions: (b) vorticity snapshot at UTC: 2023-12-31 21:00, (c) spatial spectrum, and (d) temporal spectra at select locations shown in (a). }
    \label{fig:modelcompare}
\end{figure}

\paragraph{Experimental protocol}
We demonstrated the potential of our framework by comparing three models: a deterministic mean-field regression, an end-to-end diffusion, and our TAUDiff model, each with $O(10)$M trainable parameters overall. For our training, we use the atmospheric reanalysis dataset (ERA5) at $0.25^\circ$ lat-lon resolution produced by the European Center for Medium-range Weather Forecasts (ECMWF)~\cite{hersbach2020era5}. Instead of using coarse interpolation, we use lowpass spherical wavelet filtering~\cite{schroder1995spherical, da2020local} to create band-limited low-resolution ERA5 fields to ensure proper scale separation. This approach closely mirrors real-world scenarios where bias-corrected GCM data lacks fine-scale spatio-temporal features. The models were trained over 40 years of ERA5 atmospheric wind data over Europe (1980-2020) (figure~\ref{fig:modelcompare}(a)), and validated over 2021-23. All the models were trained on a single T4 GPU over 50 epochs, with training times of 24, 48, and 60 hours for the mean, end-to-end diffusion, and the TAUDiff models, respectively. 
\paragraph{Validation} Diffusion models can create noisy artifacts in the samples and they only get magnified when plotting the gradients of the outputs. Hence, for a rigorous qualitative evaluation, we consider vorticity snapshots instead of using wind speeds like in other works~\cite{merizzi2024wind}. Qualitatively, the vorticity contour predictions of mean and TAUDiff models demonstrate dynamical consistency of storm fronts, whereas the end-to-end diffusion model distorts them due to noise injection (circled zone in figure~\ref{fig:modelcompare}(b)). Quantitatively, pointwise statistics computed over validation years 2021-23 show good recovery of spatial and temporal spectrum for TAUDiff, while mean model underrepresents, and end-to-end diffusion overrepresents higher temporal frequencies, respectively (figures~\ref{fig:modelcompare}(c) and \ref{fig:modelcompare}(d)).

\begin{figure}[!htbp]
    \centering
    \includegraphics[width=\linewidth]{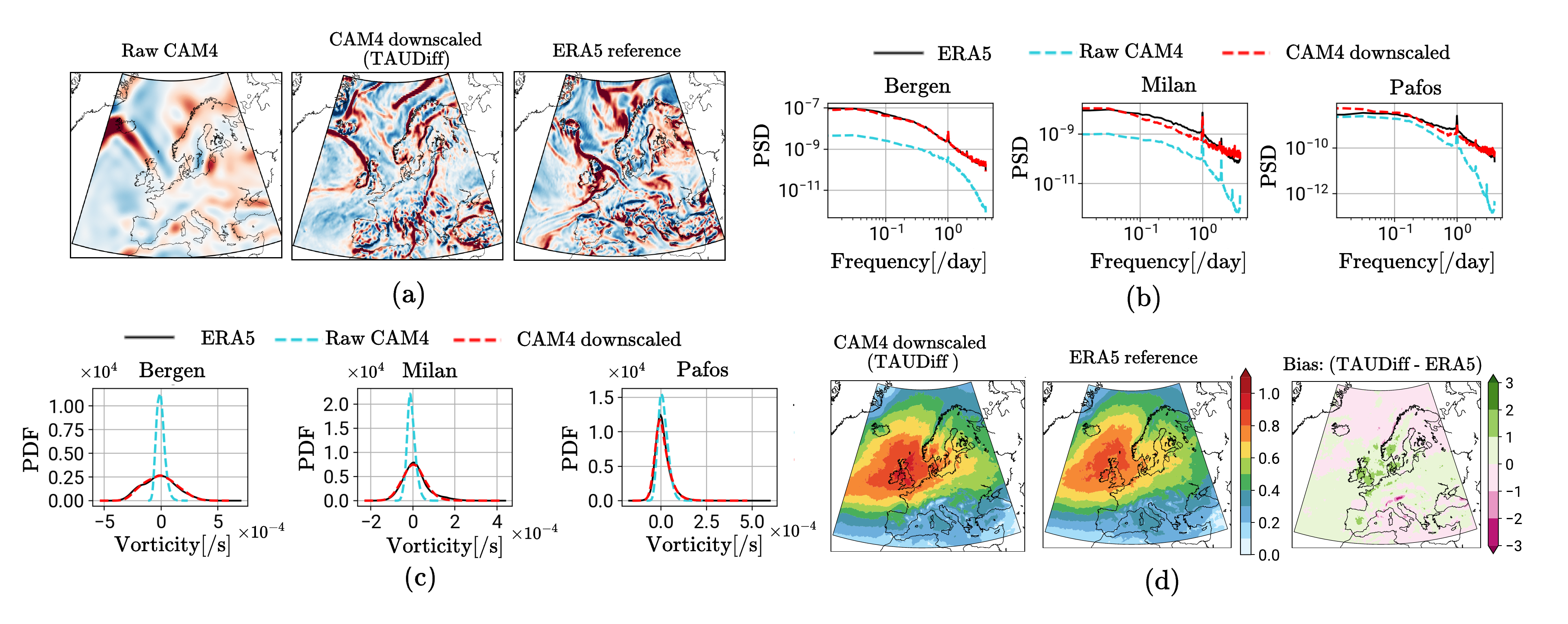}
    \caption{Assessment of downscaling performance on bias corrected CAM4 data: (a) Vorticity contours at a representative time instance, (b) temporal spectrum, (c) vorticity distributions, and (d) local storm counts.}
    \label{fig:cam4downscale}
\end{figure}

\paragraph{Testing}
The performance of our TAUDiff model was then evaluated on downscaling bias-corrected coarse GCM obtained wind fields over 40 years. We use the Community Atmosphere Model 4.0 (CAM4)~\cite{neale2010description} (at $1^\circ$ resolution) as the coarse GCM in this study. Bias correction is done by quantile-mapping~\cite{maraun2013bias} the 40-year distribution of each grid cell to that of ERA5, wavelet-filtered to GCM resolution.
As earlier, we obtain physically consistent output, and remarkable spectral recovery (figures~\ref{fig:cam4downscale}(a-c)). Although only a simple quantile mapping~\cite{maraun2013bias} is adopted for bias correction, we see good agreement with ERA5 ground truth in the local storm counts (see figure~\ref{fig:cam4downscale}(d)). 

This cements the need for a stochastic correction using a diffusion model for accurate extreme-event risk estimation. 
\subsection{Extension to km-scale downscaling}
\begin{figure}[!htbp]
    \centering
    \includegraphics[clip=True, trim={0 0 5cm 0},width=0.85\linewidth]{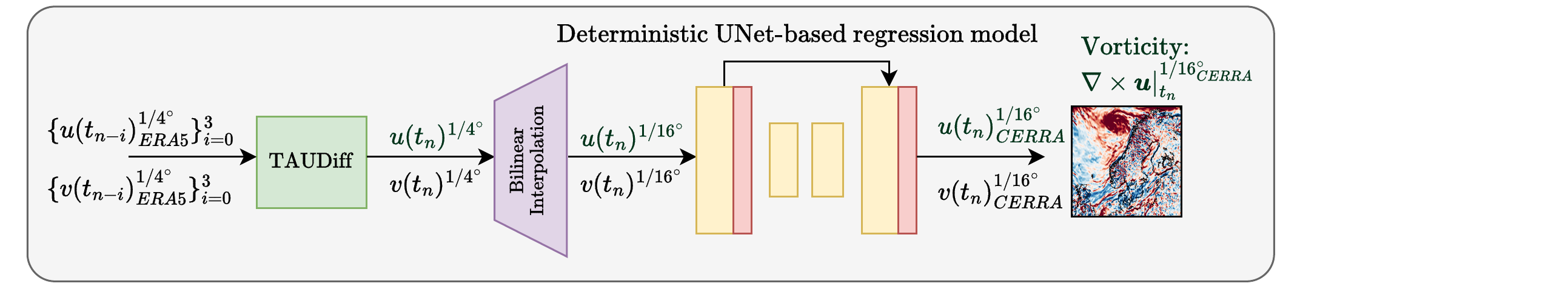}
    \caption{Schematic of the km-scale downscaling pipeline.}
    \label{fig:cerramodel}
\end{figure}

\begin{figure}[!t]
    \centering
    \includegraphics[width=\linewidth]{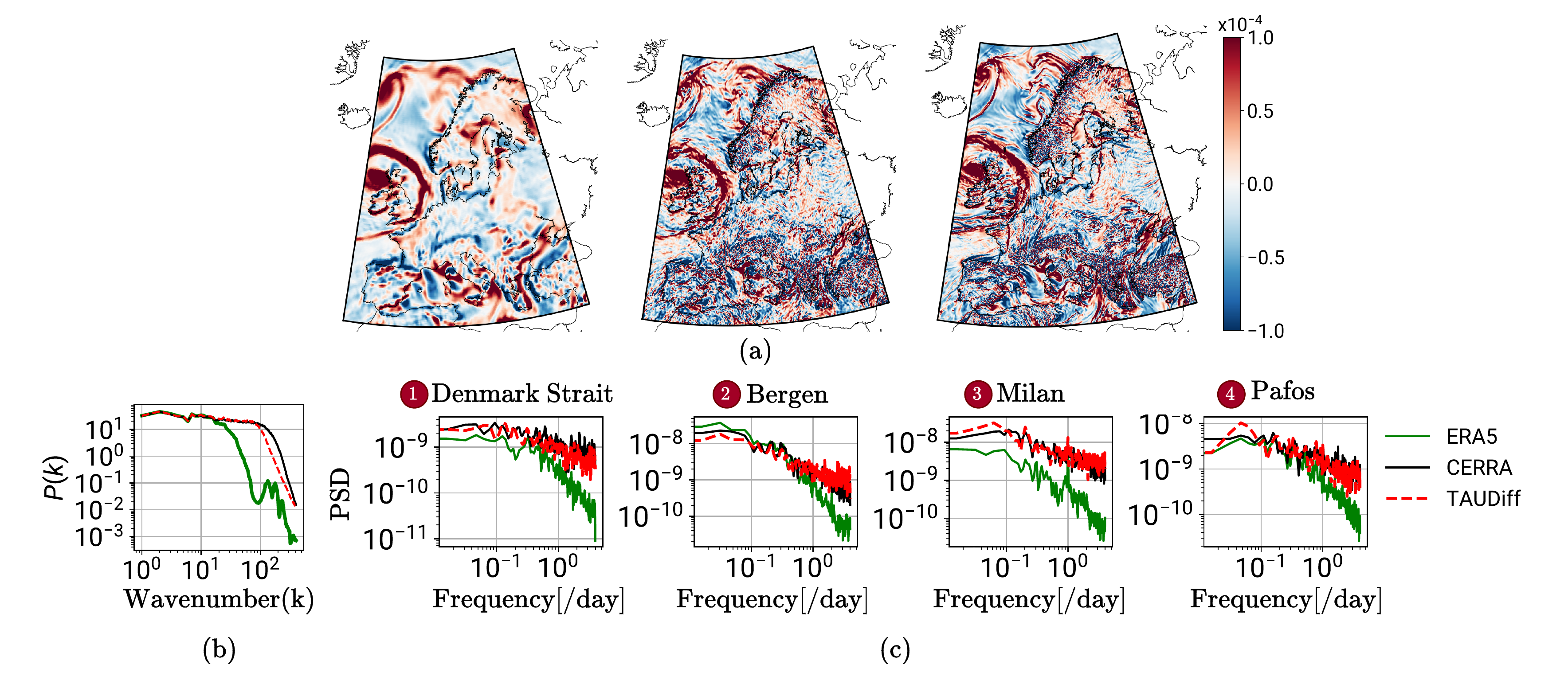}
    \caption{Assessment of ERA5 to CERRA downscaling performance: (a) Vorticity contours at UTC: 2010-11-10 21:00, (b) spatial spectrum, and (c) temporal spectra at select locations as shown in ~\ref{fig:modelcompare}(a).}
    \label{fig:cerradownscale}
\end{figure}

\paragraph{Methodology extension}
Diffusion models require multiple function evaluations while sampling~\cite{song2020score, song2020denoising}. As a result, in case of km-scale regional downscaling, ensemble methods~\cite{merizzi2024wind}, a large model size~\cite{karras2022elucidating}, and high image resolutions can vastly increase the inference times. Hence, for applications necessitating km-scale downscaling, it would be beneficial to have TAUDiff operate at a coarser resolution to reduce inference times. Since the models can be trained on reanalysis data, a single ensemble member of the diffusion model should be representative of the field-statistics. The generated samples at coarser resolution can then be downscaled using deterministic UNet based regression model to recover the fine-resolution data as depicted in the schematic figure~\ref{fig:cerramodel}.
\paragraph{Experimental protocol}
As a proof of concept, we consider the example of downscaling ERA5 atmospheric wind velocity fields at $0.25^{\circ}$ resolution to the Copernicus European Regional Reanalysis (CERRA)~\cite{schimanke2021cerra} dataset resolution of $0.0625^{\circ}$. The CERRA dataset is natively obtained on a cartesian grid. However, in our study, we project the CERRA data onto a lat-lon grid of similar resolution. Unlike Merrizi~{\it et al.}~\cite{merizzi2024wind}, where only a sub-region encompassing Italy and Alps was considered, we consider the entire European region for training as shown in figure~\ref{fig:modelcompare}(a).
While the size of mean model component of TAUDiff remains the same as in the earlier experiments, the correction-diffusion, and the deterministic UNet based regression models now consist $O(10)$M, and $O(10^2)$ thousand trainable parameters, respectively. This is to ensure that the finer scales are well captured by the models. We train this TAUDiff model over 10 years (2011-2020) of input-target pairs of ERA5, and $0.25^{\circ}$ interpolated CERRA over the European region (see figure~\ref{fig:modelcompare}(a)), and tested over the year 2010. The deterministic UNet based regression model was independently trained over the same domain using high resolution CERRA wind velocity fields as targets, and interpolated CERRA wind velocity fields as the model inputs. At inference, we chain TAUDiff, and the regression model together to generate a sample. 
\paragraph{Testing}
We obtain physically consistent fields with good qualitative, and quantitative agreement with CERRA data (see figure~\ref{fig:cerradownscale}). If one were to pass inputs at $0.0625^{\circ}$ resolution to the diffusion correction model, it can take approximately 76 minutes for downscaling one year on a single NVIDIA H100 GPU. However, since TAUDiff is now operating at $0.25^{\circ}$ resolution, we obtain a reasonable inference time of approximately $4$ minutes per one year of data. In both the cases, 20 reverse diffusion steps were considered. With frameworks like NVIDIA TensorRT, our preliminary investigations indicate that it is also possible to further reduce the inference times by up to three times the original. 

\section{Conclusion}
Overall, our proposed video-prediction-based TAUDiff model and its km-scale downscaling extension demonstrates dynamically consistent downscaling, remarkable reconstruction of spatio-temporal fine scale features, and viable inference times with the use of a small correction diffusion model. Since coarse and fine scale content of the atmospheric fields are resolved well, accurate estimation of storm statistics was possible and excellent performance on spectrum and storm statistics were obtained. It was also demonstrated as a proof of concept that even when TAUDiff operated on coarser resolutions, the ERA5-CERRA downscaling performance was remarkable. Thus, we show that TAUDiff has immense potential in accurate, and computationally efficient estimation of extreme weather events. Our future work would involve staging of TAUDiff models to obtain multi-resolution outputs for extreme weather event simulations while maintaining reasonable inference times. 

{
\small
\bibliography{workshopbiblio}
}


\end{document}